\pdfoutput=1
\documentclass{article}

\usepackage[preprint]{log_2026}
\usepackage[numbers,compress,sort]{natbib}
\usepackage{booktabs}
\usepackage{graphicx}
\usepackage{wrapfig}
\usepackage{amssymb}
\usepackage{amsthm}
\usepackage{algorithm}
\usepackage{algorithmic}
\usepackage{tikz}
\usetikzlibrary{arrows.meta,positioning,calc}
\usepackage{fontawesome5}

\DeclareMathOperator*{\argmin}{arg\,min}
\DeclareMathOperator*{\argmax}{arg\,max}
\DeclareMathOperator{\Tr}{Tr}

\title{Towards Adversarially Robust Multigraph Neural Networks for Financial
    Crime Analysis}

\author[D. Heijmans et al.]{%
    \textbf{Dennis Heijmans%
    \thanks{These authors contributed equally to this work.}
    \quad H.~{\c C}a\u{g}r\i{} Bilgi\footnotemark[1]
    \quad Rui Wang
    \quad Kubilay Atasu}\\[4pt]
    Delft University of Technology\\
    Delft, The Netherlands\\
    \texttt{\{dheijmans, cbilgi, r.wang-8, katasu\}@tudelft.nl}
}

\begin{document}

\maketitle

\begin{abstract}
    Financial transaction networks are naturally represented as multigraphs,
    where repeated transactions between the same pair of accounts are modeled as
    distinct edges. Recently proposed multigraph neural network architectures
    leverage this structure explicitly to achieve a strong performance on
    anti-money laundering (AML) tasks. However, their robustness to adversaries
    that manipulate transaction structures to evade detection remains largely
    unexplored. We present the first scalable framework for structural attacks
    and adversarial training of multigraph neural networks using node-edge
    incidence matrices, adapting projected gradient descent and projected
    randomized block coordinate descent methods to optimize structural
    perturbations over such matrices. We further introduce a cycle-based
    regularizer that penalizes changes in the global counts of directed 2- and
    3-cycles, providing a structural measure of attack plausibility. Experiments
    on synthetic pattern-detection and realistic AML datasets show that our
    learned attacks substantially outperform non-learnable perturbation
    baselines in reducing detection performance (PR-AUC), while adversarial
    training consistently improves robustness against these attacks. These
    results demonstrate that multigraph neural networks are vulnerable to
    learned structural attacks and that adversarial training provides an
    effective defense. Our code is available at
    \url{https://github.com/denheijmans/robust_multignn}.
\end{abstract}

\suppressfloats[t]
\begin{figure}[t]
    \hspace{-0.45in}
    \mbox{\begingroup
\colorlet{caught}{red!70!black}
\colorlet{restored}{green!55!black}
\colorlet{evaded}{black!50}
\colorlet{cyclecol}{teal!70!black}
\colorlet{gnncol}{blue!55!black}
\colorlet{moneycol}{green!45!black}

\newcommand{\gnodes}[1]{%
    \node[gfraud] (#1A) at (0,1.05)      {\faMask};
    \node[gb]     (#1B) at (1.18,0.36)   {\faUniversity};
    \node[gp]     (#1C) at (0.50,-1.00)  {\faUser};
    \node[gb]     (#1D) at (-0.98,-0.68) {\faUniversity};
    \node[gp]     (#1E) at (-1.15,0.52)  {\faUser};
}

\newcommand{\gedgesorig}[1]{%
    \draw[ge] (#1A) to[bend left=22]  (#1B);
    \draw[ge] (#1A) to[bend left=4]   (#1B);
    \draw[ge] (#1A) to[bend right=14] (#1B);
    \draw[ge] (#1B) to (#1C);
    \draw[ge] (#1C) to[bend left=10] (#1A);
    \draw[ge] (#1C) to[bend left=12] (#1D);
    \draw[ge] (#1C) to[bend right=8] (#1D);
    \draw[ge] (#1D) to (#1E);
    \draw[ge] (#1A) to (#1E);
}

\newcommand{\gedgespert}[1]{%
    \draw[ge]      (#1A) to[bend left=22]  (#1B);
    \draw[ge]      (#1A) to[bend left=4]   (#1B);
    \draw[gremoved] (#1A) to[bend right=14] (#1B);
    \draw[gadded]   (#1A) to[bend right=10] (#1D);
    \draw[ge] (#1B) to (#1C);
    \draw[ge] (#1C) to[bend left=10] (#1A);
    \draw[ge] (#1C) to[bend left=12] (#1D);
    \draw[ge] (#1C) to[bend right=8] (#1D);
    \draw[ge] (#1D) to (#1E);
    \draw[ge] (#1A) to (#1E);
}

\resizebox{\textwidth}{!}{%
    \begin{tikzpicture}[
            font=\sffamily,
            gp/.style     = {circle, draw=black!60, fill=blue!8,   minimum size=0.52cm,
                    inner sep=0pt, font=\footnotesize, text=black!75},
            gb/.style     = {circle, draw=black!60, fill=orange!20, minimum size=0.52cm,
                    inner sep=0pt, font=\footnotesize, text=black!75},
            gfraud/.style = {circle, draw=black!85, fill=black!88, minimum size=0.52cm,
                    inner sep=0pt, font=\footnotesize, text=white},
            ge/.style       = {-{Stealth[length=1.3mm]}, line width=0.4pt, color=black!58},
            gremoved/.style = {-{Stealth[length=1.3mm]}, line width=0.5pt, color=black!38, dotted},
            gadded/.style   = {-{Stealth[length=1.3mm]}, line width=0.6pt, color=caught, dashed},
            title/.style = {font=\sffamily\bfseries\small, text=black!90, anchor=north},
            capt/.style  = {font=\sffamily\scriptsize, text=black, align=center,
                    text width=3.9cm, anchor=north},
            flag/.style  = {circle, draw=caught, line width=1pt, minimum size=0.74cm},
            scan/.style  = {rounded corners=4pt, line width=0.7pt,
                    dash pattern=on 2.2pt off 1.6pt, minimum width=3.3cm,
                    minimum height=3.3cm},
            sgtag/.style = {font=\sffamily\scriptsize\bfseries, fill=white, inner sep=1.5pt},
            dol/.style   = {font=\tiny},
            pillok/.style = {draw=restored, fill=restored!12, text=restored, rounded corners=2pt,
                    font=\sffamily\scriptsize\bfseries, inner sep=2.5pt},
            pillno/.style = {draw=caught, fill=caught!10, text=caught, rounded corners=2pt,
                    font=\sffamily\scriptsize\bfseries, inner sep=2.5pt},
        ]

        \def\sx{2.0}\def\tx{6.6}\def\ux{11.2}\def\vx{15.8}

        \foreach \a/\b in {\sx/\tx, \tx/\ux, \ux/\vx}{%
                \draw[-{Stealth[length=3mm]}, line width=1.5pt, color=black]
                (\a+1.78,0.05) -- (\b-1.78,0.05);
            }

        \begin{scope}[shift={(\sx,0)}]
            \node[title] at (0,2.35) {1.~Transactions};
            \gnodes{a}
            \gedgesorig{a}
            \node[dol, text=caught]   at (0.75,1.07)  {\faDollarSign\faDollarSign\faDollarSign};
            \node[dol, text=caught]   at (1.00,-0.30) {\faDollarSign};
            \node[dol, text=caught]   at (0.26,0.04)  {\faDollarSign};
            \node[dol, text=moneycol] at (-0.20,-0.61){\faDollarSign\faDollarSign};
            \node[dol, text=moneycol] at (-1.24,-0.07){\faDollarSign};
            \node[dol, text=moneycol] at (-0.58,0.97) {\faDollarSign};
            \node[capt] at (0,-1.7) {People and banks send repeated payments; parallel
                edges between the same pair make it a \textbf{multigraph}.};
        \end{scope}

        \begin{scope}[shift={(\tx,0)}]
            \node[title] at (0,2.35) {2.~GNN detection};
            \gnodes{b}
            \gedgesorig{b}
            \node[scan, draw=gnncol] (bscan) at (0,0.05) {};
            \node[sgtag, text=gnncol] at (bscan.north) {\raisebox{-0.14em}{\small\faProjectDiagram}\;GNN};
            \node[flag] at (bA) {};
            \node[font=\scriptsize, text=caught] at ($(bA)+(0.46,0.30)$) {\faExclamationTriangle};
            \node[pillok] at (0,-1.6) {Detected~\faCheck};
            \node[capt] at (0,-1.95) {A \textbf{GNN}, an AI model trained on the network,
                flags suspicious accounts.};
        \end{scope}

        \begin{scope}[shift={(\ux,0)}]
            \node[title] at (0,2.35) {3.~Adversarial attack};
            \gnodes{c}
            \gedgespert{c}
            \node[scan, draw=caught] (cscan) at (0,0.05) {};
            \node[sgtag, text=caught] at (cscan.north) {\raisebox{-0.14em}{\small\faUserSecret}\;GNN + perturbation};
            \node[font=\small, text=cyclecol] at (0.48,0.32) {\faSync};
            \node[font=\sffamily\tiny\bfseries, text=cyclecol, align=center]
            at (0.50,-0.10) {cycle dist.\\preserved};
            \node[pillno] at (0,-1.6) {Evaded~\faTimes};
            \node[capt] at (0,-1.95) {The attacker rewires transactions to \textbf{evade}
                detection; the \textbf{short-cycle distribution} stays statistically
                unchanged.};
        \end{scope}

        \begin{scope}[shift={(\vx,0)}]
            \node[title] at (0,2.35) {4.~Adversarial training};
            \gnodes{d}
            \gedgespert{d}
            \node[scan, draw=restored] (dscan) at (0,0.05) {};
            \node[sgtag, text=restored] at (dscan.north) {\raisebox{-0.14em}{\small\faIcon{shield-alt}}\;robust GNN};
            \node[flag] at (dA) {};
            \node[font=\scriptsize, text=caught] at ($(dA)+(0.46,0.30)$) {\faExclamationTriangle};
            \node[pillok] at (0,-1.6) {Detected~\faCheck};
            \node[capt] at (0,-1.95) {Retraining on these attacks \textbf{hardens} the GNN
                and \textbf{restores} detection.};
        \end{scope}
    \end{tikzpicture}%
}%
\endgroup}
    \caption{Financial transactions form a multigraph; a GNN flags suspicious
        accounts; an adversary rewires transactions to evade detection while a
        cycle plausibility term keeps the short-cycle distribution statistically
        unchanged; and adversarial training restores robustness against the
        attacks.}
    \label{fig:teaser}
\end{figure}
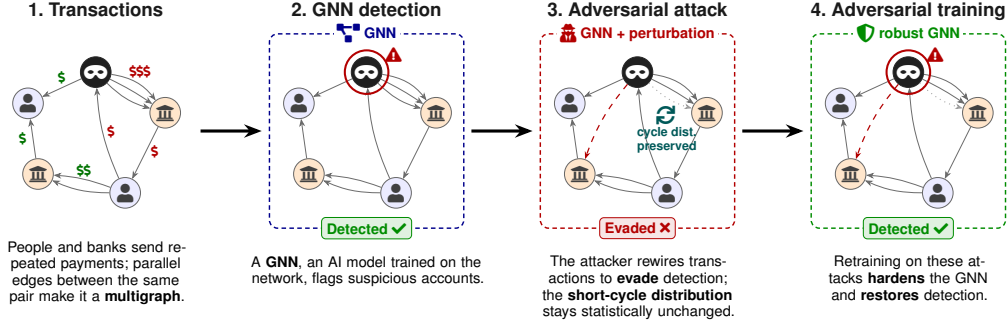

\section{Introduction}
Money laundering involves transferring proceeds of crime through the financial
system, with an estimated 2--5\% of global GDP laundered each
year~\cite{unodc_illicit_2011}. Illegal transfers are hidden within billions of
transactions flowing between legitimate accounts every day. Detecting them
increasingly relies on automated analysis of the transaction networks formed by
those payments. Graph neural networks (GNNs) have emerged as the leading tool
for this task~\cite{altman_realistic_2023,egressy_provably_2024,%
lin_fraudgt_2024,sotiropoulos_adamm_2023,ding_effective_2024}, reaching high
detection accuracy in structural pattern analysis and transaction
classification.

Furthermore, financial crime is often structured to evade
detection~\cite{jensen_searching_2025}. Breaking large transfers into smaller
ones, routing funds through intermediary accounts, and timing transactions to
fall below reporting thresholds are documented evasion tactics that reshape the
financial transaction graph. This threat raises a concrete question: can an
adversary manipulate the structure of a transaction network by adding or
removing transactions to cause a graph neural network to miss criminal activity?

Importantly, financial transaction networks are naturally modeled as
multigraphs, where the same pair of accounts may exchange multiple payments over
time and each transaction has its own attributes, such as amount and timestamp.
These attributes are essential for capturing financial crime patterns in
transaction graphs. Consequently, a family of
methods~\cite{egressy_provably_2024,lin_fraudgt_2024,bilgi_mega_2026},
henceforth referred to as multigraph neural networks, has been specifically
developed to handle multigraphs with attributed edges. Unlike other
approaches~\cite{sotiropoulos_adamm_2023,ding_effective_2024}, these methods
construct explicit representations for each multigraph edge. The prevalence of
parallel transactions further motivates the multigraph representation. In the
evaluated AML datasets, more than half of the occupied directed account pairs
contain multiple transactions, and these pairs account for over 91\% of all
transactions~\citep{bilgi_mega_2026}. Consequently, preserving individual
parallel edges is essential for accurately representing financial transaction
networks.

A substantial body of work has studied adversarial attacks and defenses for
GNNs. Structural attacks~\cite{zugner_adversarial_2018,zugner_adversarial_2019,%
xu_topology_2019,geisler_robustness_2021} generate topology perturbations by
greedily inserting or deleting edges, differentiating through bilevel
optimization objectives, or relaxing adjacency matrix entries to enable
projected first-order and randomized block-coordinate optimization. Defenses
mitigate such perturbations by reweighting or pruning suspicious
edges~\cite{zhang_gnnguard_2020}, certifying predictions under bounded
structural perturbations~\cite{wang_certified_2021}, or adversarially training
models on perturbed graphs~\cite{xu_topology_2019,gosch_adversarial_2023}.
Despite their methodological differences, these approaches fundamentally rely on
adjacency matrix representations of simple graphs. However, this assumption does
not extend naturally to multigraphs, where multiple parallel edges may connect
the same pair of nodes. Importantly, despite the growing adoption of multigraph
neural networks, their adversarial robustness has so far remained largely
unexplored.

We close this gap by introducing the first adversarial attack and defense
framework designed for multigraph neural networks, illustrated in
Figure~\ref{fig:teaser}. Our contributions can be summarized as follows.

\textbf{Structural attacks on multigraph GNNs.} We present the first
gradient-based structural attack framework for multigraph neural networks.
Existing methods differentiate with respect to adjacency matrix entries, which
cannot represent parallel edges. We instead reformulate multigraph neural
network computations using node--edge incidence matrices, which preserve edge
multiplicity while making the attack objective differentiable with respect to
the graph structure. We derive this reformulation for multigraph variants of
GIN~\cite{xu_how_2019} and PNA~\cite{corso_principal_2020}, and this
reformulation can be applied to any multigraph neural network that can be
written as a differentiable function of the incidence matrix.

\textbf{Attack plausibility.} We introduce a loss term that penalizes changes in
the global counts of directed 2- and 3-cycles under attack. These cycles
correspond to reciprocal transfers and triangular routing patterns that underlie
common laundering schemes. Therefore, an attack that distorts their counts is
detectable by a statistical monitor tracking these counts. Constraining the
perturbation to preserve them means evasion is not paid for with a visible
change in these statistics, following the principle that realistic perturbations
must be within the original graph distribution~\cite{jo_measuring_2025}.

\textbf{Scalable attacks and adversarial defenses.} We scale our attack and
defense framework to large financial networks by adapting projected randomized
block coordinate descent to the incidence-matrix setting, so memory no longer
grows with the size of the network. We further show that adversarial training
hardens Multi-GNNs against the proposed structural attacks within the same
framework.

\section{Background and Related Work}

GNNs learn representations of relational data by aggregating information from
local neighborhoods~\cite{kipf_semi-supervised_2017}. This aggregation is
commonly expressed either through adjacency-matrix multiplication or message
passing over an explicit edge multiset, where the former is the special case of
the latter. In adjacency-based formulations, node representations
$\boldsymbol{H} \in \mathbb{R}^{n\times d}$ are updated through multiplications
with the adjacency matrix $\boldsymbol{A} \in \{0,1\}^{n\times n}$, where $n$ is
the number of graph nodes, as in GIN~\cite{xu_how_2019}:
\begin{equation}
    \boldsymbol{H}^{(k+1)}=\mathrm{MLP}^{(k)}\!\left(
    (1+\epsilon^{(k)})\boldsymbol{H}^{(k)}+
    \boldsymbol{A}\boldsymbol{H}^{(k)}\right).
\end{equation}
PNA~\cite{corso_principal_2020} generalizes this sum aggregation by combining
several aggregators (mean, minimum, maximum, and standard deviation) with
degree-based scalers.

Financial transaction networks are naturally represented as directed,
edge-attributed multigraphs: accounts correspond to nodes, while each
transaction is represented as an individual edge carrying its own features.
Message-passing GNN formulations~\cite{gilmer_neural_2017,%
battaglia_relational_2018} can operate directly on explicit edge
representations, and therefore, be applied to such multigraphs. However, more
specialized architectures have also been proposed~\citep{egressy_provably_2024,%
lin_fraudgt_2024}. Notably, Multi-GNN~\citep{egressy_provably_2024} extends
standard message-passing GNNs with adaptations, such as ego identifiers, port
numbering, and reverse message passing, to obtain a more expressive GNN for
multigraphs, and FraudGT~\cite{lin_fraudgt_2024} reuses the same adaptations.

Structural attacks on GNNs typically formulate perturbations over adjacency
matrices. Nettack and Metattack~\cite{zugner_adversarial_2018,%
zugner_adversarial_2019} established the standard threat model, where an
adversary modifies a budgeted number of adjacency entries and optimizes the
resulting perturbation through gradients of the task loss. Subsequent work
improves scalability~\cite{xu_topology_2019,geisler_robustness_2021}, develops
adversarial training strategies~\cite{gosch_adversarial_2023}, and emphasize
that realistic attacks should remain plausible and difficult to
detect~\cite{jo_measuring_2025}. Although the aforementioned attack and defense
mechanisms are effective for GNNs operating on simple graphs, their perturbation
space does not directly extend to multigraphs. Modifying an edge while
preserving other parallel edges cannot be represented as a perturbation of the
adjacency matrix. Attacks on multigraph GNNs therefore require a perturbation
space defined over node--edge incidence matrices.

Adversarial robustness has also been studied for hypergraph and heterogeneous
graph neural networks, which represent graph structure using node--hyperedge
incidence matrices or relation-specific adjacency matrices,
respectively~\cite{hu_hyperattack_2023,he_hypergraph_2025,cai_hypernear_2025,%
he_transferable_2026,shang_transferable_2023,zhao_hgattack_2024,wang_heta_2025}.
In hypergraph neural networks, each column of the incidence matrix represents a
hyperedge and may contain an arbitrary number of incident nodes, such that
modifying an entry amounts to adding or removing a node from a hyperedge. In
contrast, attacks against heterogeneous graph neural networks typically leverage
relation-specific gradients, relation-aware meta-paths, or sequential attacks
across different relation types. Our incidence matrix representation of
multigraphs with edge attributes differs fundamentally from both settings: each
column represents an edge between a source and a destination node, repeated
edges between the same two nodes are stored in additional columns, and a
distinct representation is computed for each edge. Moreover, our construction
does not require predefined relation types. Because existing attacks for
hypergraph and heterogeneous graph neural networks do not support repeated
hyperedges or repeated edges of the same type between the same pair of nodes,
they cannot be meaningfully applied to our setting.

\section{Method}

\subsection{Problem Statement and Threat Model}

Consider a directed, unweighted multigraph
$\mathcal{G}=(\mathcal{V},\mathcal{E})$, where $\mathcal{V}=\{v_1,\dots,v_n\}$
is a set of $n$ nodes and $\mathcal{E}=\{\!\{e_1,\dots,e_m\}\!\}$ is a multiset
of $m$ directed edges. Each edge $e_j=(u_j,w_j)$ consists of a source node
$u_j\in\mathcal{V}$ and a target node $w_j\in\mathcal{V}$. Let $f_\theta$ be a
multigraph GNN and let $\boldsymbol{y}$ denote the ground-truth labels. We
consider a white-box evasion setting in which the attacker knows the
architecture and parameters of $f_\theta$, the input multigraph $\mathcal{G}$,
and the labels $\boldsymbol{y}$. The parameters $\theta$ remain fixed during the
attack. The attacker can only modify $\mathcal{G}$ by rewiring any of its edges,
up to a budget of $\Delta\in\mathbb{N}$. Although real-world adversaries may not
have such knowledge or control, this worst-case setting yields a conservative
estimate of robustness against weaker attackers.

A structural perturbation rewires a subset of edges by replacing their target
nodes while keeping source nodes unchanged. Rewiring keeps the number of edges
fixed, so the perturbation is a fixed-dimensional variable amenable to
first-order optimization over the incidence matrix; it also matches the
anti-money laundering setting, where an adversary reroutes outgoing transactions
rather than changing the sender of incoming ones. The adversarial attack and
defense problems are defined respectively by Eq.~\eqref{eq:adv_attack} and
Eq.~\eqref{eq:adv_defense}, where $\mathcal{B}(\mathcal{G},\Delta)$ denotes the
set of all multigraphs obtainable from $\mathcal{G}$ by rewiring at most
$\Delta$ edge targets, and the goal of adversarial training is to seek model
parameters robust to worst-case perturbations within the same budget:
\begin{align}
    \max_{\widetilde{\mathcal{G}}\in\mathcal{B}(\mathcal{G},\Delta)}
    \mathcal{L}_{0/1}\bigl(f_\theta(\widetilde{\mathcal{G}}),
    \boldsymbol{y}\bigr),
    \label{eq:adv_attack} \\
    \argmin_{\theta}\max_{\widetilde{\mathcal{G}}\in
                        \mathcal{B}(\mathcal{G},\Delta)}
    \mathcal{L}_{0/1}\bigl(f_\theta(\widetilde{\mathcal{G}}),
    \boldsymbol{y}\bigr).
    \label{eq:adv_defense}
\end{align}

\subsection{Incidence-Based Multi-GNN Reformulation}
\label{sec:multi_gin}
We reformulate the message passing of multigraph
GNNs~\cite{egressy_provably_2024} using the oriented incidence matrix
$\boldsymbol{B}\in\mathbb{R}^{n\times m}$, whose rows correspond to nodes and
whose columns correspond to edges. The entry $\boldsymbol{B}_{ij}$ is $-1$ if
edge $e_j$ leaves node $v_i$, $+1$ if $e_j$ enters $v_i$, and $0$ otherwise.
With one column assigned to each edge, parallel edges are naturally
distinguished and the computation becomes expressible as matrix products rather
than an iteration over the edge multiset. The original edge-list representation
does not support differentiation with respect to the graph structure, which
gradient-based structural attacks require. To enable such attacks, we decompose
$\boldsymbol{B}$ into its positive and negative parts:
$\boldsymbol{B}_{\text{dst}}\in\{0,1\}^{n\times m}$, where
$(\boldsymbol{B}_{\text{dst}})_{ij}=1$ if $e_j$ enters $v_i$, and
$\boldsymbol{B}_{\text{src}}\in\{0,1\}^{n\times m}$, where
$(\boldsymbol{B}_{\text{src}})_{ij}=1$ if $e_j$ leaves $v_i$ (both zero
otherwise), so
$\boldsymbol{B}=\boldsymbol{B}_{\text{dst}}-\boldsymbol{B}_{\text{src}}$. We
derive this reformulation for the multigraph variants of GIN and PNA,
\emph{Multi-GIN} and \emph{Multi-PNA}.

At each layer $k$, a forward path computes a message for every edge from its
sender, receiver, and edge features and aggregates the messages at the
destination nodes. A reverse path does the same in the opposite direction. Both
node updates are averaged and combined with a skip connection. Edge features are
updated from both endpoint embeddings and the previous edge features, with a
skip connection. The message passing update at layer $k$ is defined as
\begin{equation}
    \begin{gathered}
        \boldsymbol{M}_{\text{fwd}}^{(k)}
            =\text{MESSAGE}_{\text{fwd}}^{(k)}\!\left(
            \boldsymbol{B}_{\text{src}}^{\top}\boldsymbol{H}^{(k)},\,
            \boldsymbol{B}_{\text{dst}}^{\top}\boldsymbol{H}^{(k)},\,
            \boldsymbol{Z}^{(k)}\right)\text{,}\\
        \boldsymbol{H}_{\text{fwd}}^{(k)}
            =\text{UPDATE}_{\text{fwd}}^{(k)}\!\left(
            \boldsymbol{H}^{(k)},\,\text{AGGREGATE}\!\left(
            \boldsymbol{B}_{\text{dst}},
            \boldsymbol{M}_{\text{fwd}}^{(k)}\right)\right)\text{,}\\
        \boldsymbol{M}_{\text{rev}}^{(k)}
            =\text{MESSAGE}_{\text{rev}}^{(k)}\!\left(
            \boldsymbol{B}_{\text{dst}}^{\top}\boldsymbol{H}^{(k)},\,
            \boldsymbol{B}_{\text{src}}^{\top}\boldsymbol{H}^{(k)},\,
            \boldsymbol{Z}^{(k)}\right)\text{,}\\
        \boldsymbol{H}_{\text{rev}}^{(k)}
            =\text{UPDATE}_{\text{rev}}^{(k)}\!\left(
            \boldsymbol{H}^{(k)},\,\text{AGGREGATE}\!\left(
            \boldsymbol{B}_{\text{src}},
            \boldsymbol{M}_{\text{rev}}^{(k)}\right)\right)\text{,}\\
        \boldsymbol{H}^{(k+1)}
            =\tfrac{1}{2}\!\left(\boldsymbol{H}^{(k)}+\phi\!\left(
            \text{BATCH\_NORM}\!\left(\tfrac{\boldsymbol{H}_{\text{fwd}}^{(k)}+
            \boldsymbol{H}_{\text{rev}}^{(k)}}{2}\right)\right)\right)\text{,}\\
        \boldsymbol{Z}^{(k+1)}
            =\boldsymbol{Z}^{(k)}+\tfrac{1}{2}\,
            \text{MLP}_{\text{edge}}^{(k)}\!\left[
            \boldsymbol{B}_{\text{src}}^{\top}\boldsymbol{H}^{(k+1)},\,
            \boldsymbol{B}_{\text{dst}}^{\top}\boldsymbol{H}^{(k+1)},\,
            \boldsymbol{Z}^{(k)}\right]\text{,}
    \end{gathered}
\end{equation}
where $\boldsymbol{H}^{(k)}\in\mathbb{R}^{n\times d}$ are node features and
$\boldsymbol{Z}^{(k)}\in\mathbb{R}^{m\times d}$ are edge features at layer $k$,
$\text{MESSAGE}$ computes edge messages from the sender, receiver, and edge
features, $\phi$ is ReLU, and $\text{MLP}_{\text{edge}}^{(k)}$ updates edge
features from concatenated neighboring node embeddings and the current edge
embedding.
$\text{MESSAGE}(\boldsymbol{X}_{\text{snd}},\boldsymbol{X}_{\text{rcv}},
\boldsymbol{Z})$,
$\text{AGGREGATE}(\boldsymbol{B},\boldsymbol{M})$, and
$\text{UPDATE}(\boldsymbol{H},\boldsymbol{A})$ are defined as follows:
\begin{equation}
    \renewcommand{\arraystretch}{1.2}\setlength{\arraycolsep}{10pt}
    \begin{array}{lll}
        \toprule
                         & \text{Multi-GIN} & \text{Multi-PNA} \\
        \midrule
        \text{MESSAGE}   & \phi(\boldsymbol{X}_{\text{snd}}+\boldsymbol{Z})
                         & \text{MLP}_{\text{pre}}[\boldsymbol{X}_{\text{snd}},
                           \boldsymbol{X}_{\text{rcv}},\boldsymbol{Z}] \\
        \text{AGGREGATE} & \boldsymbol{B}\boldsymbol{M}
                         & \bigl[\,s(\boldsymbol{B}\mathbf{1})\odot
                           a_{\boldsymbol{B}}(\boldsymbol{M})\,\bigr]_{
                           s\in\mathcal{S},\,a\in\mathcal{A}} \\
        \text{UPDATE}    & \text{MLP}\bigl((1+\epsilon)\boldsymbol{H}
                           +\boldsymbol{A}\bigr)
                         & \text{MLP}_{\text{post}}[\boldsymbol{H},
                           \boldsymbol{A}] \\
        \bottomrule
    \end{array}
\end{equation}
where $\epsilon$ is a learnable scalar,
$\mathcal{A}=\{\text{mean},\min,\max,\text{std}\}$ are aggregators weighted by
the entries of $\boldsymbol{B}$, $\mathcal{S}$ are the degree scalers (identity,
amplification, attenuation) applied to the node degrees
$\boldsymbol{B}\mathbf{1}$, and $[\cdot]$ denotes concatenation. Expressing the
update in terms of $\boldsymbol{B}_{\text{src}}$ and
$\boldsymbol{B}_{\text{dst}}$ makes node and edge embeddings differentiable with
respect to $\boldsymbol{B}$, so any loss on the node or edge embeddings
propagates gradients directly to the graph structure. The minimum and maximum
are not differentiable in $\boldsymbol{B}$. During the continuous attack
optimization we therefore replace them with a weighted log-sum-exp relaxation,
while discrete graphs use exact aggregation.

\subsection{Structural Perturbations} \label{sec:structural_attacks}
To introduce perturbations to the graph structure, we employ a continuous
relaxation of the discrete attack problem, inspired by~\cite{xu_topology_2019}.
They relax adjacency entries, which merge parallel edges into one entry per node
pair. We relax the incidence matrix $\boldsymbol{B}$ instead, so that each
transaction can be perturbed individually. Let $\boldsymbol{p}\in[0,1]^{m}$
denote a vector representing the probability of rewiring each edge, and
$\boldsymbol{R}\in[0,1]^{m\times n}$ denote a matrix assigning scores to all
nodes as potential new targets for each edge. We set
$\boldsymbol{R}_{ji}=-\infty$ if $v_{i}$ is either the source or target node of
edge $e_{j}$, ensuring that an edge cannot be rewired to its original endpoints.
Applying a row-wise softmax to $\boldsymbol{R}$ then yields a probability
distribution $\boldsymbol{Q}^{(t)}_{j}$ over all candidate target nodes for each
edge $e_{j}$. We can then combine the probability vector $\boldsymbol{p}$ with
the probability matrix $\boldsymbol{Q}$ to define a relaxed incidence matrix
$\boldsymbol{\hat{B}}$, which encodes the expected structure of the perturbed
graph. Specifically, for each edge $e_{j}$, if it is not rewired (with
probability $1-\boldsymbol{p}_{j}$), it remains connected to its original target
node $w_{j}$. If the edge is rewired (with probability $\boldsymbol{p}_{j}$), it
connects to a new target node $v_{i}$ according to the probability
$\boldsymbol{Q}_{ji}$. Formally, this relationship is given by
\begin{equation}
    \boldsymbol{\hat{B}}^{(t)}_{ij}=
    \begin{cases}
        -1 & \text{if~$v_{i}=u_{j}$,} \\
        1-\boldsymbol{p}^{(t)}_{j} & \text{if~$v_{i}=w_{j}$,} \\
        \boldsymbol{p}^{(t)}_{j}\boldsymbol{Q}^{(t)}_{ji} & \text{otherwise.}
    \end{cases}
\end{equation}
\begin{figure}
    \centering
    \begin{minipage}[t]{0.4\linewidth}
        \centering
        \includegraphics[width=\linewidth]{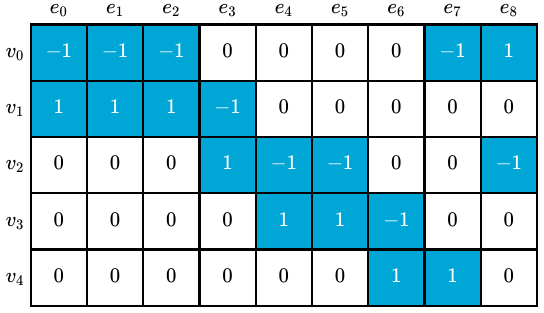}
    \end{minipage}\hspace{32pt}
    \begin{minipage}[t]{0.41\linewidth}
        \centering
        \includegraphics[width=\linewidth]{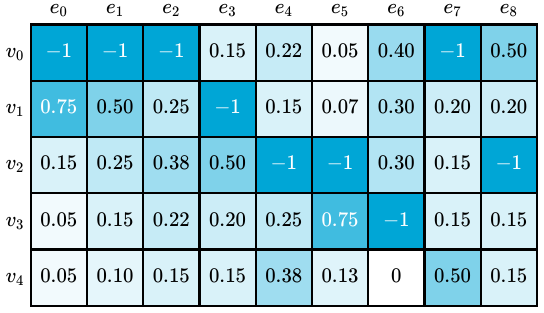}
    \end{minipage}
    \caption{The discrete oriented incidence matrix $\boldsymbol{B}$ (left) and
        its continuous relaxation $\boldsymbol{\hat{B}}$ (right).}
    \label{fig:incidence_matrices}
\end{figure}

Figure~\ref{fig:incidence_matrices} shows the discrete incidence matrix
$\boldsymbol{B}$ and its continuous relaxation $\boldsymbol{\hat{B}}$ for an
example graph. We optimize $\boldsymbol{p}$ and $\boldsymbol{R}$ jointly by
gradient descent on a differentiable surrogate loss $\mathcal{L}_\text{attack}$
(Appendix~\ref{app:surrogate_loss}), following~\cite{geisler_robustness_2021}.
After each update, we project the probability vector $\boldsymbol{p}$ onto the
feasible set, ensuring that the expected total number of rewired edges remains
within the allowed budget $\sum\boldsymbol{p}\le\Delta$ and that
$\boldsymbol{p}\in[0,1]^{m}$. We use the function $\Pi_{\mathcal{C}}$
by~\cite{ang_fast_2021} for efficiently projecting $\boldsymbol{p}$ onto the
capped simplex defined by these constraints. Formally, this projection step is
given by
\begin{equation}
    \begin{gathered}
        \boldsymbol{p}^{(t+1)}\gets\Pi_{\mathcal{C}}(\boldsymbol{p}^{(t+1)})
        \text{~where}\\\mathcal{C}=\bigl\{\boldsymbol{x}\in[0,1]^{m}\;\big|\;
        \textstyle\sum_{j=1}^{m}\boldsymbol{x}_{j}\le\Delta\bigr\}\text{.}
    \end{gathered}
\end{equation}

Finally, we generate $K$ discrete incidence matrices
$\boldsymbol{\widetilde{B}}$ from the optimized probabilities $\boldsymbol{p}$
and $\boldsymbol{Q}$ to assess the attack's effectiveness. Each matrix is
constructed by first sampling a binary vector $E^{(k)}$ that identifies which
edges are rewired, while respecting the budget constraint, and then assigning
new target nodes for the rewired edges according to $\boldsymbol{Q}$. We select
the incidence matrix that yields the lowest model PR-AUC as the final
adversarial instance. This sampling strategy is similar to that used
by~\cite{xu_topology_2019} for sampling adjacency matrices. Formally, the
sampling procedure is given by Algorithm~\ref{alg:incidence_sampling}.

\subsection{Scalable Attacks}
\label{sec:scalable_attacks}
The projected gradient descent (PGD) attack of
Section~\ref{sec:structural_attacks} performs structural perturbations on
multigraphs but scales poorly: it optimizes $\mathcal{O}(m\cdot n)$ learnable
parameters, which is infeasible in memory for real-world graphs. We therefore
adapt \emph{projected randomized block coordinate descent}
(PR-BCD)~\cite{geisler_robustness_2021} to our incidence-matrix setting.

At each iteration $t$ we operate on a restricted subset of edges
$\mathcal{E}^{(t)}\subseteq\mathcal{E}$, and for each edge
$e_{j}\in\mathcal{E}^{(t)}$ only a subset of nodes
$\mathcal{V}_{j}^{(t)}\subseteq\mathcal{V}$. Let $b_{\mathcal{E}}\ll m$ and
$b_{\mathcal{V}}\ll n$ denote the dimensions of this block, which stay fixed
regardless of graph size, reducing the learnable parameters to
$\mathcal{O}(b_{\mathcal{E}}\cdot b_{\mathcal{V}})$. The relaxed incidence
matrix $\smash{\boldsymbol{\hat{B}}}$ of Section~\ref{sec:structural_attacks} is
restricted to the active block: the $\boldsymbol{p}_{j}\boldsymbol{Q}_{ji}$ term
applies only when $e_{j}\in\mathcal{E}^{(t)}$ and
$v_{i}\in\mathcal{V}_{j}^{(t)}$, and is zero otherwise. \pagebreak
\begin{wrapfigure}[25]{r}{0.6\linewidth}
    \vspace{-12pt}
    \setlength{\columnwidth}{\linewidth}
    \begin{algorithm}[H]
        \caption{Adversarial incidence matrix selection using randomized
            perturbation sampling.}
        \label{alg:incidence_sampling}
        \begin{algorithmic}[1]
            \STATE \textbf{Input:} rewiring probabilities $\boldsymbol{p}$,
            target distributions $\boldsymbol{Q}$
            \STATE \textbf{Parameters:} rewiring budget $\Delta$,
            number of samples $K$
            \FOR{$k=1,2,\dots,K$}
            \STATE Sample binary mask indicating which edges to rewire:
            \begin{equation}
                \boldsymbol{E}^{(k)}\sim\text{Bernoulli}(\boldsymbol{p})
                \quad\text{s.t.}\quad\sum_{j=1}^{m}E^{(k)}_j\le\Delta
            \end{equation}
            \vspace{-\belowdisplayskip}
            \STATE Sample new target nodes for the selected edges:
            \begin{equation}
                \tilde{w}^{(k)}_j=
                \begin{cases}
                    w_j & \text{if~}E^{(k)}_j=0                       \\
                    v_i & \text{with~probability~}\boldsymbol{Q}_{ji}
                \end{cases}
                \quad,\forall{}j
            \end{equation}
            \vspace{-\belowdisplayskip}
            \STATE Construct the perturbed incidence matrix:
            \begin{equation}
                \widetilde{\boldsymbol{B}}^{(k)}_{ij}=
                \begin{cases}
                    -1 & \text{if~}v_i=u_j               \\
                    +1 & \text{if~}v_i=\tilde{w}^{(k)}_j \\
                    0  & \text{otherwise}
                \end{cases}
                \quad,\forall{}i,j
            \end{equation}
            \vspace{-\belowdisplayskip}
            \ENDFOR
            \STATE \textbf{Output:} Perturbed incidence matrix
            $\widetilde{\boldsymbol{B}}^{(k^*)}$ where
            \begin{equation}
                k^*=\argmax_{k}\;\mathcal{L}_{0/1}\big(
                f_{\theta}(\widetilde{\boldsymbol{B}}^{(k)},
                \boldsymbol{X},\boldsymbol{Z}),\boldsymbol{y}
                \big)
            \end{equation}
            \vspace{-\belowdisplayskip}
        \end{algorithmic}
    \end{algorithm}
\end{wrapfigure}
To keep exploring the perturbation space, we periodically refresh the active
blocks: whenever at least half of the edge-selection probabilities
$\boldsymbol{p}$ are nonzero, we replace the $50\%$ of edges with the smallest
probabilities by newly sampled edges and target candidates. This prevents
overfitting to the initial subset of edges. After a predetermined number of
iterations we stop resampling and optimize the final block, then sample discrete
incidence matrices as described above. The block size must satisfy
$b_{\mathcal{E}}\geq\Delta$ so the block holds enough edges to use the full
rewiring budget.

\subsection{Attack Plausibility}
A perturbation that degrades the GNN performance is only useful to an adversary
if it also escapes a defender inspecting the graph for signs of tampering.
Unlike the image domain, graphs offer no visual criterion, so plausibility must
be defined through numerical measures on graph
statistics~\cite{jo_measuring_2025}. To maximize plausibility, we introduce a
regularization term that constrains the perturbation to preserve the statistical
distribution of short cycles, following the cycle-counting formulas
of~\cite{harary_number_1971,voropaev_number_2012}. We focus on cycle lengths two
and three, which are tractable to count exactly and correspond to the reciprocal
transfers and triangular routing patterns that characterize common laundering
schemes.

To characterize cycle structure, we define the multigraph adjacency matrix
$\boldsymbol{A}=\boldsymbol{B}_{\text{src}}\boldsymbol{B}_{\text{dst}}^{\top}$,
where entry $\boldsymbol{A}_{ij}$ counts the number of directed edges from node
$v_{i}$ to node $v_{j}$ (proved in Appendix~\ref{app:adjacency_proof}). The
diagonal entry $(\boldsymbol{A}^{c})_{ii}$ then counts directed closed walks of
length $c$ starting at $v_{i}$, so $\Tr(\boldsymbol{A}^{c})/c$ gives the total
number of directed closed walks of length $c$ in the graph. For $c=2$ and $c=3$,
these coincide with true simple cycles. For $c\geq4$, however, walks revisit
nodes and edges, so $\Tr(\boldsymbol{A}^{c})/c$ overcounts distinct cycles. We
therefore limit the regularizer to cycle lengths up to $C=3$.

The cycle regularizer measures the normalized mean squared error in cycle counts
between the original and perturbed graph. Because the regularizer should
penalize the addition or removal of cycles rather than the exact number of
parallel edges forming them, we apply an elementwise saturation function
$\phi\colon\mathbb{R}\to[0,1]$ to the adjacency matrices before raising them to
the $c$-th power. During training we use $\phi=\tanh$, which is differentiable
and saturates to $1$ for large positive entries. For evaluation of the metric on
a discrete graph we use $\phi(x)=\min(x,1)$, which clips multi-edges to $1$
exactly but is not differentiable. The cycle regularizer is then
\begin{equation}
    \mathcal{L}_{\text{cycle}}=\frac{1}{n(C-1)}\sum_{c=2}^{C}
    \left(\frac{\Tr(\phi(\boldsymbol{A})^{c})-\Tr(\phi(\boldsymbol{A}')^{c})}{c}
    \right)^{2}\text{,}
\end{equation}
where $\boldsymbol{A}'
=\boldsymbol{B}'_{\text{src}}(\boldsymbol{B}'_{\text{dst}})^{\top}$
is the adjacency matrix of the perturbed graph. Minimizing
$\mathcal{L}_{\text{cycle}}$ keeps the directed cycle profile of the perturbed
graph close to the original. Combining the cycle regularizer with the attack
objective yields the final surrogate loss
$\mathcal{L}'=\mathcal{L}_{\text{attack}}+\lambda\mathcal{L}_{\text{cycle}}$,
where $\mathcal{L}_{\text{attack}}$ is the differentiable surrogate loss and
$\lambda\geq0$ balances attack effectiveness against cycle preservation.

\subsection{Adversarial Training}
\label{sec:adv_training}
We harden multigraph GNNs against structural perturbations with adversarial
training~\cite{madry_towards_2018,gosch_adversarial_2023}. The inner
maximization of Eq.~\eqref{eq:adv_defense} has no closed-form solution, so at
every training step we approximate it with the PR-BCD attack of
Section~\ref{sec:scalable_attacks}, generating a fresh adversarial example
within the rewiring budget $\Delta$ and updating $\theta$ on the perturbed
graph.

\section{Empirical Evaluation}
\label{sec:empirical_evaluation}

We evaluate the framework along its three contributions: the structural attack,
the cycle plausibility term, and adversarial training as a defense.
Section~\ref{sec:eval_synthetic} examines all three on a synthetic multi-pattern
detection benchmark, and Section~\ref{sec:eval_aml} investigates whether the
findings extend to large-scale, realistic AML transaction graphs.

\subsection{Synthetic Pattern-Detection Experiments}
\label{sec:eval_synthetic}

\subsubsection{Experimental Setup}
\label{sec:eval_setup}

\textbf{Dataset.} We evaluate on the synthetic multigraph pattern-detection
benchmark contributed in~\cite{egressy_provably_2024}. Each split is an
independently generated directed multigraph of $8192$ nodes with average degree
$6$, giving roughly $49{,}000$ transaction edges; training, validation, and test
graphs use separate seeds, so the setting is fully inductive. Every node carries
a binary label for each of $11$ structural patterns: in-degree, out-degree,
fan-in, fan-out, directed cycles of length two through six, scatter-gather, and
biclique. These patterns mirror money laundering typologies such as smurfing and
layering. The statistics of these patterns are provided in
Appendix~\ref{app:data_stats}.

\textbf{Metric.} Pattern labels are heavily imbalanced, so we report the area
under the precision--recall curve (PR-AUC), which summarizes the trade-off
between precision and recall over all decision thresholds and, unlike ROC-AUC,
is not inflated by the many true negatives~\cite{davis_relationship_2006,%
saito_precision-recall_2015}. We report it per pattern and as the macro average
over the $11$ patterns. Where the positive class is the majority, we report
PR-AUC for the negative class instead, so every value corresponds to the
minority class.

\textbf{Model.} We evaluate our incidence-matrix reformulation of
Multi-GIN~\cite{egressy_provably_2024}, described in
Section~\ref{sec:multi_gin}, retaining reverse message passing, edge-feature
updates, ego-node identifiers, and port numbering. All models use six
message-passing layers and are trained with neighborhood
sampling~\cite{hamilton_inductive_2017}. Full architectural and optimization
hyperparameters are given in Appendix~\ref{app:implementation_details}.

\textbf{Attack protocol.} All attacks use the PR-BCD attack of
Section~\ref{sec:structural_attacks} in the white-box evasion setting with fixed
model parameters. The rewiring budget $\Delta$ is derived from a fixed attack
ratio of the total number of edges in the graph. The non-learnable baseline,
denoted \emph{naive}, rewires the same number of edges to uniformly random
targets under the identical discrete-sampling procedure. Full attack
hyperparameters are given in Appendix~\ref{app:implementation_details}.

\textbf{Seeds.} Every reported point aggregates $9$ runs ($3$ model seeds
$\times$ $3$ attack seeds); bands denote one standard deviation. All experiments
run on a single GPU.

\subsubsection{Structural Attack Effectiveness}
\label{sec:eval_attack}

\begin{wrapfigure}[18]{r}{0.5\linewidth}
    \vspace{-\dimexpr\intextsep+\baselineskip+4.8pt\relax}
    \centering
    \includegraphics[width=0.9\linewidth]{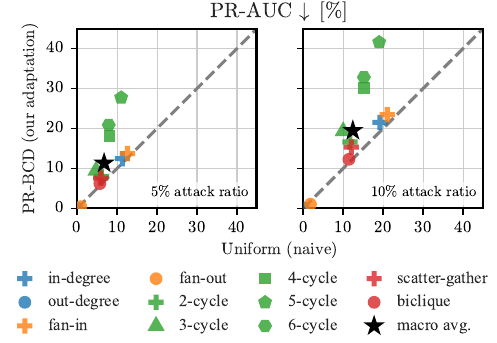}
    \caption{PR-AUC drop (percentage points) per pattern, comparing the learned
        PR-BCD attack ($y$-axis) with uniform random rewiring ($x$-axis), at a
        $5\%$ (left) and $10\%$ (right) rewiring budget. Each marker is one of
        the $11$ structural patterns, with the macro average shown as a star.
        The dashed line is the diagonal $y=x$.}
    \label{fig:attack_scatter}
    \vspace{-2\intextsep}
\end{wrapfigure}

We first compare the learned PR-BCD attack against the non-learnable naive
baseline at equal budget. Figure~\ref{fig:attack_scatter} plots the PR-AUC drop
of each attack per pattern. Almost every pattern, and the macro average, lies
above the diagonal at both the $5\%$ and $10\%$ attack ratios, so the learned
attack removes more detection PR-AUC performance than uniform random rewiring.
The gap is largest for the higher-order cycle patterns, which the learned attack
disturbs while random rewiring rarely reaches.

We assess the robustness of the undefended multigraph GIN under PR-BCD at
increasing attack ratios. Figure~\ref{fig:robustness_budget} traces PR-AUC per
pattern across the sweep. The degradation is strongly uneven: detection of $5$-
and $6$-cycles falls from near-perfect to below $60\%$ at a $10\%$ attack ratio,
whereas biclique and scatter-gather remain much closer to their clean PR-AUC\@.
Out-degree and fan-out, omitted from the figure, are near-invariant by
construction: the attack rewires only edge targets, leaving out-degree unchanged
and fan-out nearly so.

\suppressfloats[t]
\begin{figure}[t]
    \begin{minipage}[t]{0.42\linewidth}
        \centering
        \includegraphics[width=\linewidth]{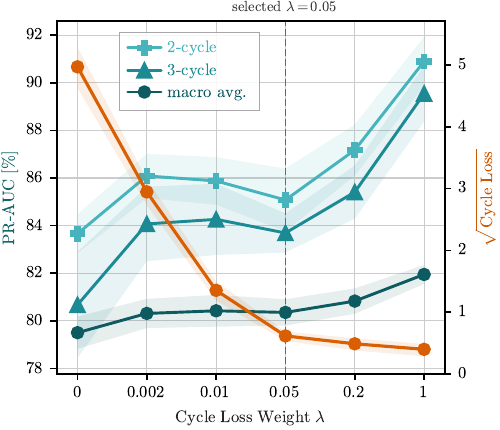}
        \caption{Attack effectiveness (measured by PR-AUC, left) against
            macro-level plausibility (measured by the square root of cycle-count
            error, right) as the plausibility weight $\lambda$ varies, at a
            $10\%$ rewiring budget. Bands show standard deviation ($9$ runs).}
        \label{fig:lambda_tradeoff}
    \end{minipage}\hfill
    \begin{minipage}[t]{0.54\linewidth}
        \centering
        \includegraphics[width=\linewidth]{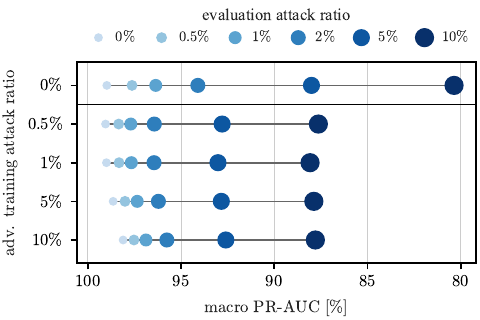}
        \caption{Macro PR-AUC of adversarially trained models (rows; $0\%$ is
            the undefended baseline), each evaluated under PR-BCD at test
            budgets from $0\%$ to $10\%$. Dot color and size encode the test
            attack ratio, and the $x$-axis is reversed. Attacks use
            $\lambda=0.05$; Appendix~\ref{app:ablation_cycle} repeats this with
            $\lambda=0$.}
        \label{fig:adv_training}
    \end{minipage}
\end{figure}

\subsubsection{Plausibility Trade-offs}
\label{sec:eval_cycle}

To quantify the impact of cycle preservation on attack effectiveness, we sweep
the plausibility weight $\lambda$ at an attack rate of $10\%$.
Figure~\ref{fig:lambda_tradeoff} reports attack effectiveness, measured by
PR-AUC, and plausibility, measured by the square root of the difference between
the cycle counts of the original and perturbed graphs, considering only directed
cycles of lengths two and three. Preserving these statistics reduces the
structural discrepancy between the original and perturbed graphs, making the
attacks more plausible. However, such preservation does not guarantee that the
perturbations remain undetectable by mechanisms that rely on other
graph-structural or transaction-level statistics.

Raising $\lambda$ from $0$ to $0.05$ significantly reduces the cycle-count loss,
while macro PR-AUC under attack increases by less than one point. Beyond
$\lambda = 0.05$ the cycle-count loss decreases only marginally, yet the attack
success rate decreases considerably: PR-AUC for the $2$-cycles and $3$-cycles as
well as the macro-average across all the tasks rises steadily as $\lambda$ grows
toward $1$. In summary, at $\lambda=0.05$, the attacks remain both plausible and
effective, and increasing $\lambda$ further offers no tangible benefit.

\subsubsection{Adversarial Training Defense}
\label{sec:eval_defense}

To test whether adversarial training restores robustness to the structural
attack, we train models with the procedure of Section~\ref{sec:adv_training} at
several budgets and evaluate each under the PR-BCD attack, with the plausibility
weight fixed at $\lambda=0.05$, the trade-off point identified in
Section~\ref{sec:eval_cycle}. Figure~\ref{fig:adv_training} reports macro PR-AUC
across test budgets from $0\%$ to $10\%$. The undefended baseline falls from
$99\%$ on the clean graph to $80\%$ under a $10\%$ attack. Every adversarially
trained model instead retains about $88\%$ under the same attack while keeping
clean PR-AUC near $99\%$. The recovered robustness is largely insensitive to the
attack ratio used during adversarial training: training against a $0.5\%$ attack
ratio already yields most of the benefit. These gains are not an artifact of the
plausibility term: with $\lambda=0$ during both the adversarial training of the
models and their robustness evaluation, the recovery is undiminished (see
Appendix~\ref{app:ablation_cycle} for further details).

\subsection{AML Experiments}
\label{sec:eval_aml}

\subsubsection{Experimental Setup}
We evaluate the proposed structural attack and adversarial training defense on
the realistic IBM AML benchmark~\cite{altman_realistic_2023}. We consider two
dataset variants: Small with a higher-illicit (HI) and lower-illicit (LI) ratio
setting. Dataset statistics are given in Appendix~\ref{app:data_stats}. Accounts
are represented as nodes and transactions as directed edges, and the task is to
classify each transaction as licit or illicit. We report PR-AUC with the illicit
class as the positive class, computed by ranking all transactions of the
evaluated split by their predicted probability of being illicit. Following the
standard protocol for this benchmark~\cite{egressy_provably_2024}, we adopt a
temporal inductive split where transactions are partitioned chronologically.

We use the same model as in Section~\ref{sec:eval_setup} with three
modifications: two message-passing layers instead of six, a binary
edge-classification head in place of the node multilabel head, and a
class-weighted cross-entropy loss to counter the extreme class imbalance. The
threat model is defined in Section~\ref{sec:eval_threat_model_aml}, and the
attack and adversarial training protocol in Section~\ref{sec:eval_protocol_aml}.
The cycle regularizer is not used in the AML experiments. Remaining
implementation details are in Appendix~\ref{app:implementation_details}.

\subsubsection{Threat Model}
\label{sec:eval_threat_model_aml}

We consider an attacker who launders money through a set of accounts. We define
\emph{criminal accounts} as accounts that send or receive at least one illicit
transaction in the graph snapshot, in any time period, including accounts whose
first illicit transaction falls in the evaluated period. The attacker can rewire
its illicit transactions to any criminal account, modeling the rerouting of
illicit funds through accounts associated with laundering activity. We assume a
white-box attacker: it knows the trained model, including its parameters, the
transaction graph on which the model makes predictions, and which transactions
and accounts belong to the laundering activity. These constraints make our
attack more representative of AML scenarios than unconstrained structural
attacks, although the white-box assumption remains a worst-case simplification.
We also report results under the unconstrained threat model in
Appendix~\ref{app:aml_unconstrained}

\textbf{Plausible perturbations.} The attacker can rewire illicit transactions
only. Rewiring a transaction replaces its receiver account with another criminal
account, while leaving the sender account and all transaction attributes
unchanged. These constraints preserve the transaction-level characteristics of
existing transfers while restricting structural changes to illicit activity.
Licit transactions are never modified, and no transactions are added or removed.
For an attack ratio $\rho$, the attacker rewires exactly $\Delta$ transactions,
a fraction $\rho$ of the illicit transactions of the evaluated split. Thus, the
attack budget is defined relative to the number of illicit transactions rather
than the total number of transactions. Rewired transactions retain their illicit
labels because they continue to represent transfers between criminal accounts.
Since rewiring is restricted to criminal accounts, the labels of licit
transactions remain unchanged. However, predictions for licit transactions may
still change because their sampled neighborhoods can include rewired
transactions.

\subsubsection{Adversarial Attacks and Defenses}
\label{sec:eval_protocol_aml}

We use the PR-BCD attack restricted to the plausible perturbations. The attack
maximizes the tanh-margin loss~\citep{geisler_robustness_2021} on the illicit
transactions. After optimization, we take the rewiring probabilities of the
epoch with the highest attack loss, draw $50$ discrete perturbations from them,
each rewiring exactly $\Delta$ transactions, and keep the one that yields the
lowest PR-AUC on the evaluated split. As a non-learnable baseline, uniform
random rewiring selects $\Delta$ illicit transactions uniformly at random and
rewires each of them to a criminal account drawn uniformly at random. It obeys
the same constraints as PR-BCD and uses the same selection of the best of $50$
sampled perturbations. We evaluate attack ratios
$\rho\in\{0,1,2,5,10,20,50\}\%$. Each point is the mean over $3$ attack seeds
against a single trained model per dataset, and bands denote one standard
deviation over these attack seeds.

Adversarial training follows Section~\ref{sec:adv_training} under the threat
model above. The adversarially trained model is initialized from the trained
standard model and fine-tuned. In every epoch, PR-BCD perturbs the training
graph at an attack ratio of $20\%$, rewiring illicit transactions of the
training split among the criminal accounts of the training snapshot, and the
model is trained on the perturbed graph. After each epoch, the validation graph
is attacked in the same way, and we keep the checkpoint with the highest PR-AUC
on the perturbed validation split.

\subsubsection{Structural Attack Effectiveness and Adversarial Training on AML
Graphs}

\begin{figure}[tb]
    \centering
    \includegraphics[width=0.85\linewidth]{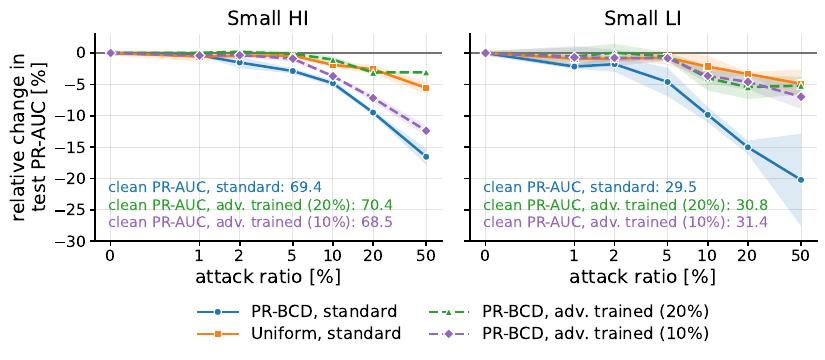}
    \caption{Relative change in test PR-AUC on Small HI and Small LI under our
        threat model as the attack ratio increases. Each line shows the change
        relative to the same model's clean (un-attacked) PR-AUC, which is listed
        in each panel; $0$ means no degradation. \emph{Standard} is the
        undefended model; \emph{adv.\ trained ($x\%$)} is the model fine-tuned
        against PR-BCD at an attack ratio of $x\%$.}
    \label{fig:aml_small}
\end{figure}

Figure~\ref{fig:aml_small} reports the relative change in test PR-AUC on Small
HI and Small LI as the attack ratio increases from $0\%$ to $50\%$. PR-BCD
degrades the standard model more than uniform random rewiring at every attack
ratio of $2\%$ or more, and the gap widens with the budget. Under the plausible
criminal threat model, the attacker can only rewire its illicit transactions
among accounts it controls, without adding accounts or transactions or changing
transaction attributes. Despite these restrictions, PR-BCD reduces PR-AUC by up
to one fifth, indicating that targeted structural changes can substantially
impair detection. In contrast, uniform random rewiring under the same
constraints causes much smaller drops, suggesting that vulnerability depends on
which transactions are rewired and where they are redirected, rather than on
rewiring alone. The impact also depends on the attack budget: for $\rho \leq
5\%$, PR-AUC drops by less than $5\%$ on both datasets. This suggests that
substantially degrading detection requires the attacker to rewire a larger share
of its illicit transactions, increasing the extent of structural changes needed
to evade detection. Appendix~\ref{app:aml_pna} reports the same attack against
Multi-PNA.

Adversarial training with PR-BCD at $\rho=20\%$ substantially improves
robustness. Even at an attack ratio of $\rho=50\%$, the PR-AUC drop is
comparable to that caused by uniform random rewiring on the undefended model.
Moreover, increasing the attack ratio from the training level of $20\%$ to
$50\%$ does not further increase the PR-AUC drop. The adversarial training
attack ratio also matters. On Small HI, the model adversarially trained at
$\rho=20\%$ suffers a smaller PR-AUC drop than the model trained at $\rho=10\%$
at every tested attack ratio. For the latter model, the drop continues to
increase as the attack ratio exceeds $10\%$. Finally, adversarial training has
little effect on clean performance: clean PR-AUC increases for models trained at
$\rho=20\%$ and decreases slightly for the model trained at $\rho=10\%$, with
all changes limited to $1.3$ percentage points.

\section{Conclusion and Discussion}
The incidence-matrix reformulation of multigraph GNN message passing, which we
derive for Multi-GIN and Multi-PNA, enables the first gradient-based structural
attack on multigraph GNNs. PR-BCD attacks substantially outperform uniform
random rewiring at the same budget, confirming that the framework exploits
structural vulnerabilities that random rewiring misses. The cycle plausibility
regularizer penalizes changes in the global counts of directed 2- and 3-cycles.
With a small regularization weight, the attack changes these counts as little as
with any larger weight, while its effect on PR-AUC remains nearly the same as
without regularization. Larger weights only weaken the attack. Because the
regularizer considers only 2- and 3-cycles, it does not guarantee that the
attack remains unnoticed by a monitor that tracks other graph statistics. Such
statistics can, however, be added to the attack objective as further
plausibility terms in the same way. Scaling via projected randomized block
coordinate descent makes the framework memory-efficient for large financial
networks, and adversarial training shows that multigraph GNN robustness can be
recovered after structural attack, enabling both evaluation and defense within a
single framework.

On the IBM AML benchmark, attacks remain effective under a plausible criminal
threat model in which money launderers can rewire only their own illicit
transactions among criminal accounts. PR-BCD reduces PR-AUC by up to 20\%,
whereas uniform random rewiring under the same constraints causes substantially
smaller drops. Adversarial training mitigates this vulnerability, keeping the
PR-AUC drop comparable to that caused by uniform random rewiring on the
undefended model, even when half of the illicit transactions are rewired. For
broader impact, discussion, and ethical considerations, please refer to
Appendix~\ref{app:broader_impact},~\ref{app:discussion},
and~\ref{app:ethical_considerations}, respectively.

\section*{Acknowledgments}
The support of the Swiss National Science Foundation (project number: 212158)
for this work is gratefully acknowledged. Research reported in this work was
partially facilitated by computational resources and support of the Delft AI
Cluster (DAIC) at TU Delft (RRID: SCR\_025091)~\citep{daic_delft_2024} but
remains the sole responsibility of the authors, not the DAIC team.

\bibliographystyle{unsrtnat}
\bibliography{references}

\appendix
\setcounter{equation}{0}
\setcounter{table}{0}
\setcounter{figure}{0}
\numberwithin{equation}{section}
\numberwithin{table}{section}
\numberwithin{figure}{section}

\newpage
\section{Proof of Adjacency Matrix Factorization}
\label{app:adjacency_proof}
\begin{proof}[Proof that $\boldsymbol{A}
    =\boldsymbol{B}_{\text{src}}\boldsymbol{B}_{\text{dst}}^{\top}$]
    Fix any pair of nodes $v_i, v_j \in \mathcal{V}$. By definition,
    $(\boldsymbol{B}_{\text{src}})_{ik}=1$ if edge $e_k$ leaves $v_i$ and $0$
    otherwise, and $(\boldsymbol{B}_{\text{dst}})_{jk}=1$ if edge $e_k$ enters
    $v_j$ and $0$ otherwise. By the definition of matrix multiplication,
    $(CD)_{ij}=\sum_{k}C_{ik}D_{kj}$. Substituting
    $C=\boldsymbol{B}_{\text{src}}$ and $D=\boldsymbol{B}_{\text{dst}}^{\top}$,
    and applying the transpose identity
    $(\boldsymbol{B}_{\text{dst}}^{\top})_{kj}
    =(\boldsymbol{B}_{\text{dst}})_{jk}$,
    the $(i,j)$ entry of the product is
    \begin{equation}
        (\boldsymbol{B}_{\text{src}}\boldsymbol{B}_{\text{dst}}^{\top})_{ij}
        =\sum_{k=1}^{m}(\boldsymbol{B}_{\text{src}})_{ik}\,
        (\boldsymbol{B}_{\text{dst}})_{jk}\text{.}
    \end{equation}
    Because both factors are $\{0,1\}$-valued, the $k$-th term equals $1$ if and
    only if $e_k$ both leaves $v_i$ and enters $v_j$, i.e., $e_k=(v_i,v_j)$. The
    sum therefore counts the number of directed edges from $v_i$ to $v_j$, which
    is $\boldsymbol{A}_{ij}$ by definition. Since $(i,j)$ was arbitrary,
    $\boldsymbol{A}
    =\boldsymbol{B}_{\text{src}}\boldsymbol{B}_{\text{dst}}^{\top}$.
\end{proof}

\section{Multilabel Surrogate Loss}
\label{app:surrogate_loss}
The accuracy-based loss $\mathcal{L}_{0/1}$ in the attack objective is
non-differentiable and cannot be optimized directly with gradient-based methods.
Following~\cite{geisler_robustness_2021}, we replace it with a differentiable
surrogate. Those surrogate losses assume mutually exclusive classes and do not
generalize to multilabel settings, such as the benchmarks
of~\cite{egressy_provably_2024}, where nodes carry multiple labels. We therefore
introduce a new surrogate that handles multilabel outputs directly.

Let $z_{ic}\in\mathbb{R}$ be the raw logit for node $i\in\{1,\dots,N\}$ and
class $c\in\{1,\dots,C\}$, and let $y_{ic}\in\{0,1\}$ be the ground-truth label.
We map each binary label to a signed value $2y_{ic}-1\in\{-1,+1\}$ and define
the attack loss as
\begin{equation}
    \mathcal{L}_{\text{attack}}=\frac{1}{NC}\sum_{i=1}^{N}\sum_{c=1}^{C}
    \sigma\!\bigl((2y_{ic}-1)\,z_{ic}\bigr)\text{,}
\end{equation}
where $\sigma$ is the sigmoid function. Correctly classified nodes produce a
positive argument $(2y_{ic}-1)z_{ic}>0$, pushing $\sigma(\cdot)$ toward one.
Misclassified nodes produce a negative argument, pushing $\sigma(\cdot)$ toward
zero. In both extremes the gradient of $\sigma$ is near zero, so confidently
classified or confidently misclassified nodes contribute little to the gradient.
Gradient signal therefore concentrates on nodes with logits near zero, i.e.,
near the decision boundary, where small structural perturbations are most likely
to flip predictions. The attack thus allocates its perturbation budget where it
is most effective.

\section{Implementation Details}
\label{app:implementation_details}

\subsection{Synthetic Pattern-Detection}
\paragraph{Model hyperparameters.} Hidden dimension is $64$ and ego embedding
dimension $32$, with dropout $0.1$ and learnable $\epsilon_{\text{fwd}}$,
$\epsilon_{\text{rev}}$. Port numbers are assigned dynamically within each
mini-batch rather than globally, so the port vocabulary need only cover the
ports present in the sampled batch; port embeddings therefore use dimension $8$
over a vocabulary of $32$. Models train for $120$ epochs with Adam, learning
rate $10^{-3}$ and weight decay $10^{-4}$, under a
cosine-annealing-with-warm-restarts schedule ($T_0=8$, $T_{\mathrm{mult}}=2$,
minimum learning rate $10^{-6}$). Mini-batches are formed by neighborhood
sampling with $32$ target nodes and a per-hop fan-out of $[10,10,10,10,5,5]$ in
both edge directions.

\paragraph{Attack hyperparameters.} Each attack runs for $100$ optimization
steps with $50$ block resamples and learning rate $0.1$, selecting the strongest
of $K=25$ sampled discrete perturbations. The block size defaults to
$\sqrt{|\mathcal{V}|}$ candidate nodes and $2\Delta$ candidate edges. The
\emph{naive} baseline uses the same $K=25$ sampling procedure.

\subsection{Realistic AML Setting}
\paragraph{Model and training hyperparameters.} Hidden dimension is $32$ with
dropout $0.1$. Mini-batches contain $8{,}192$ target transactions with a per-hop
fan-out of $[50,50]$. The standard model trains with AdamW, learning rate
$3\cdot10^{-3}$ and weight decay $10^{-4}$, for at most $80$ epochs, keeping the
checkpoint with the highest validation PR-AUC (early stopping with patience $10$
after at least $20$ epochs). The positive-class loss weight is $7.0$.
Adversarial training uses the same optimizer settings for at most $20$ epochs
(patience $5$ after at least $10$ epochs).

\paragraph{Attack hyperparameters.} For evaluation, PR-BCD optimizes the
rewiring probabilities with AdamW for at most $100$ steps, stopping after $15$
steps without an increase of the attack loss (but not before step $10$), and
selects the strongest of $K=50$ sampled discrete perturbations. During
adversarial training, the attacks on the training and validation graphs use a
reduced budget of $K=20$ sampled perturbations and at most $30$ steps (patience
$8$, at least $5$ steps). On the Small datasets, the block contains every
illicit transaction of the attacked split, with every criminal account other
than the transaction's own sender and receiver as a candidate receiver. Since
the block already spans all permitted rewirings, no block resampling is
performed. On the test split, this amounts to $1{,}611$ illicit transactions and
$6{,}357$ criminal accounts for Small HI, and $802$ and $5{,}304$ for Small LI.

\paragraph{Unconstrained threat model.} The attacks in
Appendix~\ref{app:aml_unconstrained} use blocks of $512$ candidate receivers and
$400{,}000$ candidate transactions on the Small datasets and $1{,}300{,}000$ on
the Medium datasets, with $30$ block resamples for evaluation and $10$ during
adversarial training. The remaining attack hyperparameters are as above.

The positive-class loss weight is $7.0$. The reduced PR-BCD budget is $K=20$
sampled perturbations, $30$ optimization steps, and $10$ block resamples. The
attack block size is fixed per dataset scale, at $512\times400{,}000$ for the
Small datasets and $512\times1{,}300{,}000$ for the Medium datasets.

\subsection{Dataset Statistics}\label{app:data_stats}

\begin{table}[ht]
    \centering
    \caption{Positive-class ratio of each structural pattern label in the
        synthetic pattern-detection dataset.}
    \label{tab:pattern_stats}
    \begin{tabular}{lrr}
        \toprule
        Pattern        & Threshold & Positive Ratio \\
        \midrule
        In-Degree      & $>3$      & 35.5\%         \\
        Out-Degree     & $>3$      & 35.2\%         \\
        Fan-In         & $>3$      & 32.9\%         \\
        Fan-Out        & $>3$      & 32.2\%         \\
        2-Cycle        & ---       & 18.8\%         \\
        3-Cycle        & ---       & 34.4\%         \\
        4-Cycle        & ---       & 51.5\%         \\
        5-Cycle        & ---       & 67.8\%         \\
        6-Cycle        & ---       & 77.3\%         \\
        Scatter-Gather & ---       & 32.1\%         \\
        Biclique       & ---       & 31.3\%         \\
        \bottomrule
    \end{tabular}
\end{table}

\begin{table}[ht]
    \centering
    \caption{Node count, edge count, illicit-transaction rate, and
        train/validation/test split for the four IBM HI/LI datasets.}
    \label{tab:hili_stats}
    \begin{tabular}{lrrrr}
        \toprule
        Dataset       & \# Nodes  & \# Edges   & Illicit rate & Split [\%] \\
        \midrule
        AML Small HI  & 515,088   & 5,078,345  & 0.102\%      & 64/19/17   \\
        AML Small LI  & 705,907   & 6,924,049  & 0.051\%      & 64/19/17   \\
        AML Medium HI & 2,077,023 & 31,898,238 & 0.110\%      & 61/17/22   \\
        AML Medium LI & 2,032,095 & 31,251,483 & 0.051\%      & 61/17/22   \\
        \bottomrule
    \end{tabular}
\end{table}

Table~\ref{tab:pattern_stats} reports the positive-class ratio of each
structural pattern label in the synthetic pattern-detection dataset
(Section~\ref{sec:eval_synthetic}). Table~\ref{tab:hili_stats} provides the
statistics of the datasets used in Section~\ref{sec:eval_aml}.

\section{Additional Synthetic Pattern-Detection Results}

\subsection{Per-Pattern Robustness Across Attack Ratios}

\begin{figure}[ht]
    \centering
    \includegraphics[width=0.55\linewidth]{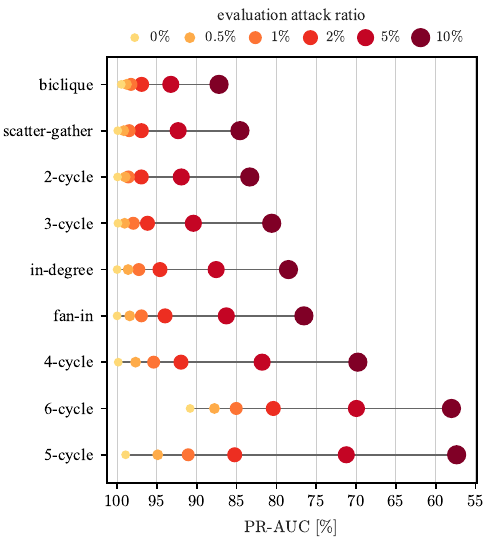}
    \caption{Per-pattern PR-AUC of the multigraph GIN under PR-BCD as the
        rewiring budget grows to $10\%$. Each of the nine rows spans its
        lowest-to-highest PR-AUC\@. Dot color and size encode the evaluation
        attack ratio, and the $x$-axis is reversed.}
    \label{fig:robustness_budget}
\end{figure}

Figure~\ref{fig:attack_scatter} of Section~\ref{sec:eval_attack} reports the
per-pattern PR-AUC drop at the $5\%$ and $10\%$ attack ratios.
Figure~\ref{fig:robustness_budget} complements that view with the absolute
PR-AUC of each pattern over the full sweep of attack ratios, from the clean
graph through $0.5\%$, $1\%$, $2\%$, $5\%$, and $10\%$, making visible both the
clean starting point of each pattern and the rate at which it degrades. The
resulting per-pattern behavior is discussed in Section~\ref{sec:eval_attack}.
Nine of the eleven patterns are shown. Out-degree and fan-out are omitted, as
destination-only rewiring leaves out-degree invariant and fan-out
near-invariant.

\subsection{Adversarial Training Without the Cycle Regularizer}
\label{app:ablation_cycle}

The synthetic adversarial-training experiment of Section~\ref{sec:eval_defense}
uses the cycle plausibility weight $\lambda=0.05$ selected in
Section~\ref{sec:eval_cycle}, applied both to the attacks that generate
adversarial examples during training and to the PR-BCD attacks used at
evaluation time. The recovered robustness reported there could therefore in
principle be attributed to the regularizer rather than to adversarial training
itself: constraining the attack to preserve the short-cycle distribution
restricts the perturbation set, and a restricted perturbation set is easier to
defend against. To rule this out, we repeat the entire experiment with
$\lambda=0$, disabling the cycle term in both the training-time and the
test-time attacks, and leave every other hyperparameter unchanged.

\begin{figure}[tbp]
    \centering
    \includegraphics[width=0.6\linewidth]{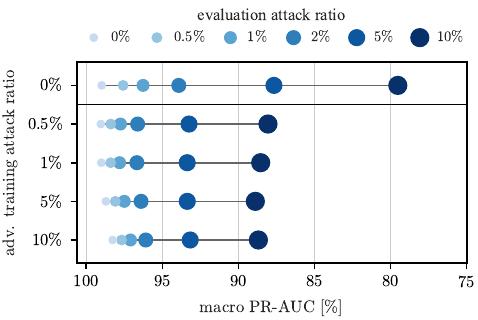}
    \caption{Macro PR-AUC of models trained with adversarial training at
        different budgets (rows; $0\%$ is the undefended baseline, separated at
        the top), each evaluated under the PR-BCD attack at test budgets from
        $0\%$ to $10\%$, with the cycle regularizer disabled ($\lambda=0$) in
        both the training-time and test-time attacks. Dot color and size encode
        the test attack ratio, and the $x$-axis is reversed. Each point averages
        $9$ runs ($3$ model seeds $\times$ $3$ attack seeds).}
    \label{fig:adv_training_no_cycle}
\end{figure}

\begin{table}[tbp]
    \centering
    \caption{Macro PR-AUC [\%] at a $10\%$ test attack budget, with and without
        the cycle plausibility term in both the training-time and test-time
        attacks. Rows are adversarial-training budgets; $0\%$ is the undefended
        baseline. Mean\,$\pm$\,std over $9$ runs ($3$ model seeds $\times$ $3$
        attack seeds).}
    \label{tab:ablation_cycle}
    \begin{tabular}{lcc}
        \toprule
        Training budget    & $\lambda=0.05$   & $\lambda=0$      \\
        \midrule
        $0\%$ (undefended) & $80.36 \pm 0.55$ & $79.52 \pm 0.71$ \\
        \addlinespace
        $0.5\%$            & $87.64 \pm 0.11$ & $88.06 \pm 0.11$ \\
        $1\%$              & $88.08 \pm 0.17$ & $88.54 \pm 0.13$ \\
        $2\%$              & $88.15 \pm 0.26$ & $88.73 \pm 0.18$ \\
        $5\%$              & $87.88 \pm 0.17$ & $88.89 \pm 0.14$ \\
        $10\%$             & $87.80 \pm 0.26$ & $88.69 \pm 0.11$ \\
        \bottomrule
    \end{tabular}
\end{table}

Figure~\ref{fig:adv_training_no_cycle} reproduces the qualitative picture of
Figure~\ref{fig:adv_training} in full: the undefended baseline degrades sharply
with the test budget, every adversarially trained model retains substantially
more macro PR-AUC under attack, and the benefit is largely insensitive to the
training budget. Table~\ref{tab:ablation_cycle} makes the comparison
quantitative at a $10\%$ test budget. The recovery is undiminished without the
regularizer and is in fact marginally larger: the best defended model improves
on the undefended baseline by $7.8$ points with $\lambda=0.05$
($80.36\rightarrow88.15$) and by $9.4$ points with $\lambda=0$
($79.52\rightarrow88.89$).

The robustness reported in Section~\ref{sec:eval_defense} is therefore
attributable to adversarial training and not to the plausibility constraint. The
cycle regularizer serves its intended purpose of keeping the perturbation
statistically unnoticeable and is not load-bearing for the defense.

\section{Additional AML Results}

\subsection{AML Results Under an Unconstrained Threat Model}
\label{app:aml_unconstrained}

In addition to the criminal threat model of Section~\ref{sec:eval_aml}, we
evaluate the attack and adversarial training under an unconstrained threat model
that allows any transaction to be rewired to any account. This broader setting
lets us assess how the model behaves under structural perturbations beyond those
permitted by the plausible criminal threat model.

\begin{figure}[ht]
    \centering
    \includegraphics[width=0.80\linewidth]{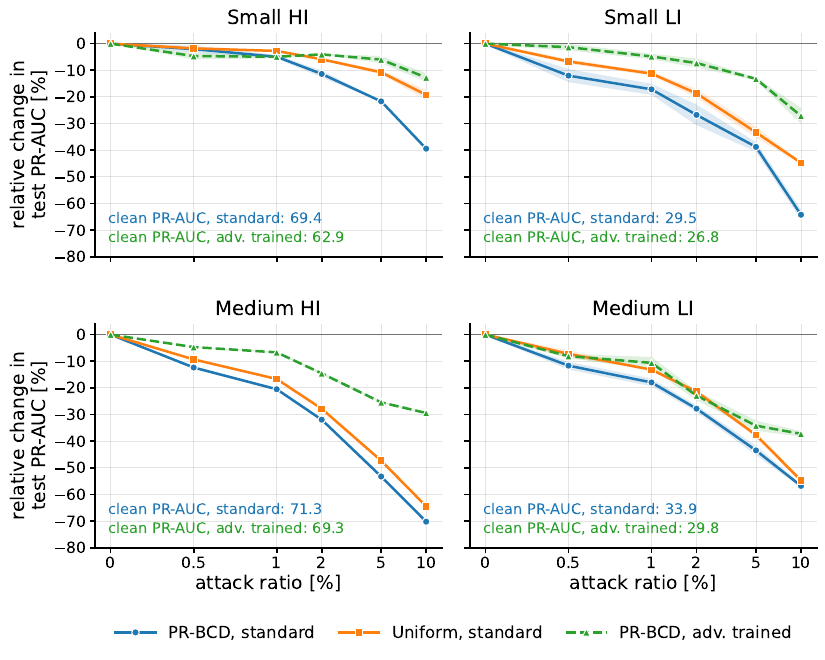}
    \caption{Relative change in test PR-AUC on the four IBM datasets under the
        unconstrained threat model as the attack ratio $\rho$ increases. Each
        line shows the change relative to the same model's clean PR-AUC, which
        is listed in each panel. \emph{Standard} is the undefended model;
        \emph{adv.\ trained} is fine-tuned against unconstrained PR-BCD at
        $\rho=5\%$. Bands show one standard deviation over attack seeds.}
    \label{fig:aml_unconstrained}
\end{figure}

\paragraph{Threat model.} As in the criminal threat model, the attacker has
white-box access to the model, the transaction graph, and the labels. Unlike
there, it may rewire \emph{any} transaction of the evaluated split, licit or
illicit, and may redirect it to \emph{any} account in the graph other than its
sender and current receiver. Rewiring again replaces only the receiving account:
the sender and all transaction attributes are unchanged, no transactions are
added or removed, and every transaction keeps its label. For an attack ratio
$\rho$, the attacker rewires a fraction $\rho$ of all transactions in the
evaluated split, including both licit and illicit transactions. This setting is
implausible because it allows the attacker to modify licit transactions between
accounts it does not control. Moreover, since illicit transactions are a small
minority, the same attack ratio corresponds to many more rewired transactions
than under the criminal threat model. Attack ratios in
Figure~\ref{fig:aml_small} and Figure~\ref{fig:aml_unconstrained} are therefore
not directly comparable.

\paragraph{Attack protocol.} PR-BCD follows Section~\ref{sec:eval_protocol_aml},
except that any transaction in the evaluated split can be rewired to any
account, and the attack maximizes cross-entropy loss over all transactions in
the split. After optimization, we sample $50$ discrete perturbations and select
the one with the lowest PR-AUC, as in the main text. Uniform rewiring randomly
selects $\Delta$ transactions from the evaluated split and redirects each to a
uniformly sampled eligible account; we apply the same best-of-$50$ selection. We
evaluate attack ratios $\rho\in{0,0.5,1,2,5,10}\%$. Each point reports the mean
over $3$ attack seeds against a single trained model per dataset. Bands indicate
one standard deviation.

\paragraph{Attack effectiveness.} Under the unconstrained threat model, PR-BCD
reduces test PR-AUC by $39\%$ to $70\%$ at $\rho=10\%$, depending on the dataset
(Figure~\ref{fig:aml_unconstrained}). Much of this damage, however, does not
require optimization. Uniform rewiring alone reduces PR-AUC by $19\%$ to $64\%$,
coming within six percentage points of PR-BCD on the Medium datasets. These
large drops show that allowing arbitrary transactions to be redirected,
including licit ones, makes the detector vulnerable to broad structural
perturbations. Since such modifications are beyond what a money launderer could
realistically carry out, these results motivate the more restrictive criminal
threat model used in the main text.

\paragraph{Adversarial training.} Adversarial training at $\rho=5\%$ improves
robustness against PR-BCD on all four datasets. At $\rho=10\%$, the largest
gains occur on Medium HI and Small LI, and the smallest on Medium LI\@. This
robustness comes at a cost in clean performance: clean PR-AUC decreases by $2.0$
to $6.5$ percentage points. On Small HI and Medium LI, the adversarially trained
model has lower PR-AUC than the standard model for attack ratios up to
$\rho=2\%$. Thus, on these datasets, the defense becomes beneficial only at
larger attack ratios.

\subsection{Adversarial Attacks on Multi-PNA}
\label{app:aml_pna}
\begin{figure}[ht]
    \centering
    \includegraphics[width=0.85\linewidth]{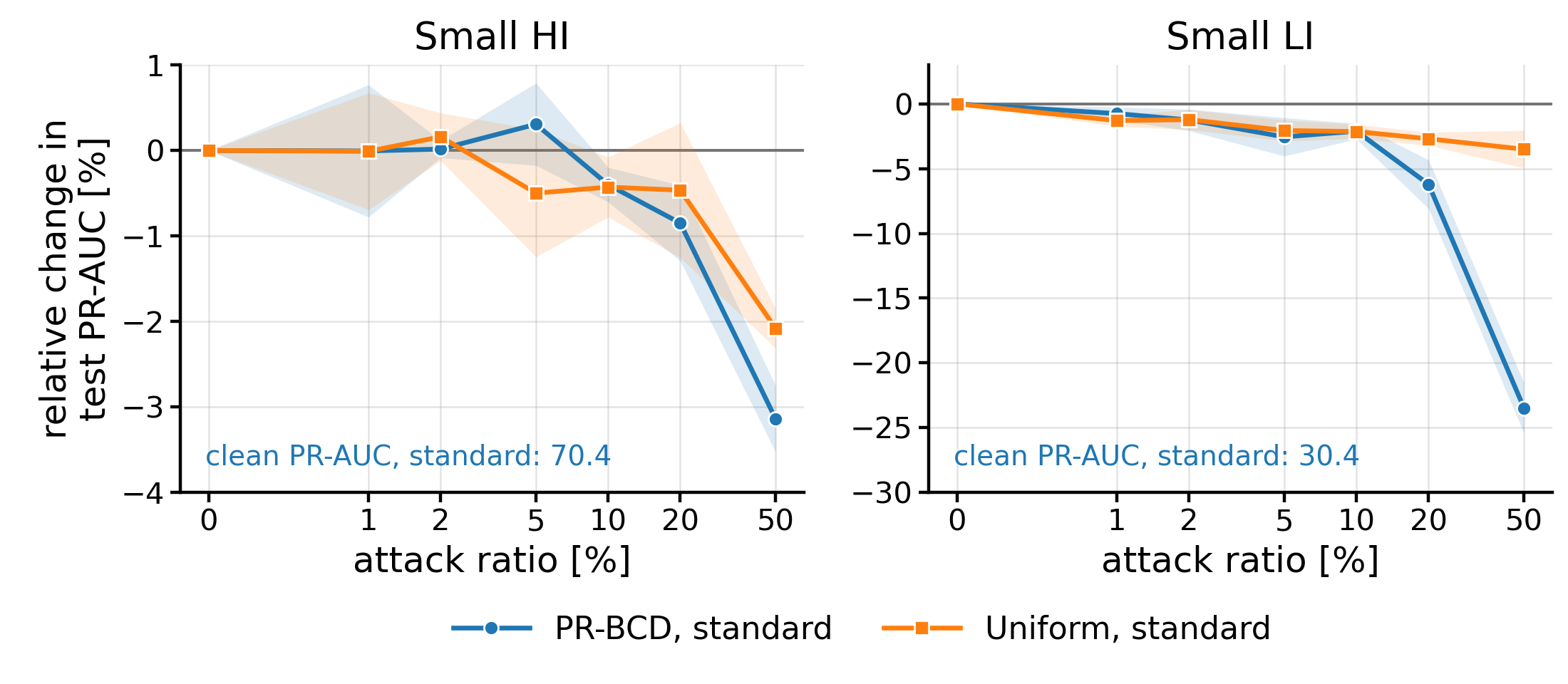}
    \caption{Relative change in test PR-AUC on Small HI and Small LI with
        Multi-PNA under the criminal threat model, as the attack ratio
        increases. Each line shows the change relative to the same model's clean
        (un-attacked) PR-AUC, which is listed in each panel; $0$ means no
        degradation. \emph{Standard} is the undefended model. Lines show the
        mean and bands one standard deviation over $3$ seeds.}
    \label{fig:aml_small_pna}
\end{figure}
Figure~\ref{fig:aml_small_pna} presents the attack results on Multi-PNA\@. On
Small HI, both PR-BCD and uniform random rewiring have a limited impact on test
PR-AUC\@. At a 50\% attack ratio, PR-BCD reduces PR-AUC by approximately 3\%,
compared to 2\% for uniform rewiring. In contrast, PR-BCD is substantially more
effective on Small LI\@. The PR-AUC drop reaches approximately 24\% at a 50\%
attack ratio, while uniform rewiring causes a drop of only 4\%. These results
prove that our proposed attack also applies to Multi-PNA.

\section{Ethical Considerations}
\label{app:ethical_considerations}
This work exposes structural vulnerabilities in multigraph neural networks used
for financial crime analysis and investigates adversarial training as a
corresponding defense. By providing a framework for worst-case robustness
evaluation, our work can contribute to the safer deployment of graph-based
anti-money laundering systems. In the following, we discuss intellectual
property, intended usage, potential misuse, risk control, and human subjects and
data privacy.

\paragraph{Intellectual property.} Our experiments rely on synthetic and
open-source datasets, established learning architectures, and existing open
research methods. We use these resources only for academic research and comply
with their applicable licenses and attribution requirements. The experiments do
not use proprietary banking records or confidential information obtained from
financial institutions.

\paragraph{Intended usage.} The proposed framework is intended for controlled
security evaluation of multigraph neural networks. Researchers, financial
institutions, and regulators may use it to identify structural failure points,
quantify model robustness under budgeted transaction rewiring, and evaluate
defensive mechanisms before deploying such models in practice. In particular,
the framework is designed to support the development of multigraph-aware
defenses and more reliable financial crime detection systems.

\paragraph{Potential misuse.} The proposed attack methods could be misused to
identify transaction-rewiring strategies that reduce the likelihood that illicit
transactions are detected. The cycle-plausibility objective may further enable
an adversary to preserve selected graph statistics while degrading model
performance, potentially facilitating the evasion of automated anti-money
laundering systems. Such misuse could result in financial losses, regulatory
violations, and the obstruction of criminal investigations. Nevertheless, the
immediate practical risk is limited by our white-box threat model and the
current scope of the framework, which considers only structural edge rewiring
rather than manipulation of transaction attributes or operational banking
infrastructure.

\paragraph{Risk control.} We conduct all experiments on synthetic and
open-source benchmark graphs and do not interact with live financial systems or
real bank transactions. Structural modifications are restricted by an explicit
perturbation budget, and our framework includes adversarial training as a
concrete defense against the proposed attacks. Any public release of the
implementation should be accompanied by clear documentation stating that it is
intended solely for authorized research, robustness assessment, and defensive
development. Practitioners should additionally combine adversarial training with
independent graph-statistics monitoring and evaluate both false negatives for
missed illicit activity and false positives affecting legitimate transactions
under clean and adversarial conditions.

\paragraph{Human subjects and data privacy.} This work does not involve human
participants or experiments using real customer records. The evaluated datasets
are synthetically generated transaction networks and do not contain personally
identifiable information. Consequently, the study does not introduce direct
privacy risks through the collection, profiling, or disclosure of individual
financial data.

\section{Discussion}
\label{app:discussion}
\subsection{Joint Structural and Feature Perturbations}
Our current attack modifies the graph structure while keeping the edge features
fixed. For AML datasets, another important direction is to perturb the edge
features and the graph structure at the same time. Edge features, such as
transaction amounts or timestamps, provide information that is complementary to
the source and destination of a transaction. It would therefore be useful to
compare three attack settings: structural perturbations only, edge feature
perturbations only, and structural and edge feature perturbations together. Such
a comparison would show whether the model is more vulnerable to changes in
transaction attributes or changes in transaction connectivity. It would also
show whether combining the two perturbation types leads to a stronger attack
than either perturbation alone. Our current framework can be extended to this
setting without changing its main optimization components. Structural
perturbations can still be optimized with the incidence-matrix formulation and
PR-BCD, while edge features can be optimized with projected gradient updates. A
simple approach is to alternate between the two updates. For example, one
optimization step can update the structural variables while keeping the edge
features fixed, and the next step can update the edge features while keeping the
structure fixed. Each update can then be projected back to its corresponding
perturbation budget. Another way is to optimize the two sets of variables
jointly. Comparing these strategies would clarify whether a more specialized
optimization method is necessary for joint structural and feature attacks. Joint
perturbations are also relevant for adversarial training. A model trained only
against structural rewiring may remain vulnerable to feature perturbations, and
the converse may also hold. Our future work could therefore evaluate adversarial
training against structural attacks, feature attacks, and their combination.

\subsection{Extending the Structural Perturbation Space}

Our current threat model preserves the source of each selected edge and rewires
only its destination. This matches the scenario in which an outgoing transaction
is rerouted to another account. However, the incidence-matrix formulation also
permits other forms of structural perturbation. First, the destination of a
transaction can be kept fixed while its source is replaced. In this way, the
attackers can modify structural properties that are unchanged by the current
destination-only rewiring. Second, both endpoints of an edge can be modified.
The attack can first keep the source fixed and use PR-BCD to select a new
destination. It can then keep the new destination fixed and optimize the source.
The order can also be reversed. This approach reuses the current optimization
procedure and does not require a separate attack algorithm. The perturbation
space could also be extended to node insertion. A new account can be represented
by adding a new row to the node--edge incidence matrix. New transaction edges
can then be added as new columns. For an inserted edge, the new node can first
be fixed as the source while PR-BCD selects a destination among existing nodes.
The new node can be fixed as the destination while PR-BCD selects a source.
Repeating these operations would allow an inserted node to participate in
several transactions. This would extend the current framework from rewiring
existing transactions to attacks that also introduce new accounts and
transactions. Although these extensions extend the space of possible attacks,
their practical relevance depends on the assumed capabilities of the adversary.
Our current destination-rewiring attack assumes that an adversary can reroute an
outgoing transaction while keeping its sender unchanged. Source rewiring instead
assumes that the sender associated with a transaction can be replaced, while
rewiring both endpoints requires control over both the sender and the receiver.
Node insertion makes a stronger assumption, as it requires the adversary to
introduce new accounts and associated transactions into the graph. Since these
assumptions represent different levels of adversarial access and may not be
equally realistic in practical AML settings, we leave their systematic
evaluation to future work.

\section{Broader Impact}
\label{app:broader_impact}
Our work introduces a systematic framework for generating structural adversarial
attacks against multigraph GNNs on large financial networks. In financial crime
detection, a successful attack means illicit transactions are classified as
legitimate, enabling money laundering schemes to evade automated systems with
direct financial and regulatory consequences. While the full-knowledge threat
model limits immediate practical risk compared to black-box settings, it
establishes an upper bound on vulnerability and provides the worst-case
evaluation that practitioners and regulators need.

The multigraph setting introduces an attack surface absent in simple graphs. An
adversary can add or reroute individual transactions between an existing pair of
nodes while keeping aggregate flow statistics, including total transaction
volume, approximately unchanged. Defenses designed for simple graphs operate on
a compressed adjacency representation and cannot detect such perturbations;
deployed multigraph detectors may therefore face a class of structural attacks
for which no existing defense was designed. This underscores the importance of
multigraph-aware robustness evaluation before deploying such systems in
practice.

Our adversarial training defense recovers the accuracy lost under attack, as
demonstrated in Section~\ref{sec:empirical_evaluation}. Nevertheless,
adversarial training introduces a robustness-accuracy trade-off: classification
accuracy on clean graphs may decrease as the model learns to resist
perturbations. Practitioners should evaluate this trade-off for their specific
deployment context and tune the plausibility constraints to match their graph's
distributional properties. Both false negative rates for missed laundering and
false positive rates for blocked legitimate transactions should be monitored
under clean and adversarial conditions.

\section{GenAI Usage Statement}
Generative AI assistants were used to support code development and to edit the
language of this paper. All research ideas, methods, experiments, and analyses
were conceived and carried out by the authors, who reviewed all AI-assisted
output and take full responsibility for the content of this paper.

\end{document}